%% file: neurips_2026.tex
\documentclass{article}

    \PassOptionsToPackage{numbers, compress}{natbib}
 \usepackage[preprint]{neurips_2026}

\usepackage[utf8]{inputenc} 
\usepackage[T1]{fontenc}    
\usepackage{hyperref}       
\usepackage{url}            
\usepackage{booktabs}       
\usepackage{amsfonts}       
\usepackage{nicefrac}       
\usepackage{microtype}      
\usepackage{xcolor}         

\usepackage[table]{xcolor}
\usepackage{graphicx}
\usepackage{wrapfig}
\usepackage{multirow}
\usepackage{amsmath}
\usepackage{amssymb}
\usepackage{bm}
\usepackage{tcolorbox}
\tcbuselibrary{raster,skins}
\usepackage[ruled,vlined]{algorithm2e}
\usepackage{pifont}
\usepackage{tikz}
\usepackage{multirow}   
\usepackage{array}
\usepackage{caption}
\usepackage{subcaption}
\usepackage{siunitx}
\usepackage{listings}
\newcommand{\cnum}[1]{\raisebox{-0.12ex}{\ding{#1}}}

\newcommand{\methodname}{GRC}
\newcommand{\tokens}{meta latent tokens}
\newcommand{\grit}{GritLM}

\hypersetup{
    colorlinks,
    linkcolor={red!50!black},
    citecolor={red!50!black},
    urlcolor={red!50!black}
}
\definecolor{softgray}{gray}{0.7}
\definecolor{myred}{RGB}{255,0,0}

\definecolor{softpink}{RGB}{248, 200, 220}
\definecolor{softblue}{RGB}{180, 210, 240}
\definecolor{bgtint}{RGB}{248, 250, 253}
\definecolor{promptpink}{RGB}{245, 200, 220}
\definecolor{promptblue}{RGB}{185, 215, 240}
\definecolor{prompttint}{RGB}{250, 251, 253}
\newtcolorbox{promptbox}[1][]{
  enhanced,
  center,
  width=0.92\linewidth,
  title=#1,
  frame style={left color=promptpink, right color=promptblue},
  colback=prompttint,
  colframe=promptblue,
  coltitle=black!85,
  boxrule=0.5pt,
  arc=2.5pt,
  left=10pt, right=10pt, top=7pt, bottom=7pt,
  before skip=8pt, after skip=8pt,
  drop fuzzy shadow=black!15
}

\title{\methodname{}: Unifying Reasoning-Driven Generation, Retrieval and Compression}

\author{%
  Zhongtao Miao\thanks{Tsuruoka Lab.} \quad Qiyu Wu\thanks{Independent Researcher.} \quad Yoshimasa Tsuruoka \\
  The University of Tokyo\\
  \texttt{\{miao, tsuruoka\}@logos.t.u-tokyo.ac.jp}
}

\begin{document}

\maketitle

\input{sections/abs}
\input{sections/intro}

\input{sections/method}
\input{sections/inference}

\input{sections/exp}

\input{sections/related}
\input{sections/conclusion}

\input{sections/author_contribution}

\bibliographystyle{unsrtnat}
\bibliography{ref}
\clearpage

\appendix
\input{sections/app}


\end{document}

%% file: sections/abs.tex
\begin{abstract}
    \emph{Text embedding} and \emph{generative tasks} are usually trained separately based on large language models (LLMs) nowadays. This causes a large amount of training cost and deployment effort.
	\emph{Context compression} is also a challenging and pressing task, which is vital to reasoning-driven generation, and agentic tasks requiring long context and continual learning.
	In this paper, we explore how to unify reasoning-driven generation, reasoning-enhanced text representation and context compression tasks in one forward pass for LLMs. 
	Through meta latent tokens and a unified generative, representative and compressive tuning approach, we propose a training framework named \methodname{} that bridges the three tasks. The trained models can  accomplish three objectives in a single forward pass while maintaining modular, LEGO-style flexibility during inference. This design greatly reduces the deployment effort for retrieval-augmented generation (RAG) and achieves efficient inference and three times data utilization during training.
	Furthermore, this framework design enables a new paradigm for text embedding: \emph{self-reason-latent embeds}, 
	and a new generation paradigm, \emph{latent memory-augmented generation}, 
    where compressed and internalized KV cache with $O(1)$ length is used as the updatable memory. 
	We also propose \emph{hybrid paged attention} to speed up the inference of our models.
	Extensive experiments on reasoning-intensive retrieval benchmarks, generative tasks, document compression, latency evaluation, and RAG settings demonstrate the effectiveness of our method and may shed light on the truly unified model that can handle reasoning-driven generation, embedding and compression tasks seamlessly\footnote{The code will be available at~\url{https://github.com/gpgg/grclm}.}. 
\end{abstract}

%% file: sections/intro.tex
\section{Introduction}
\label{sec:intro}


Large language models (LLMs)~\citep{grattafiori2024llama,yang2025qwen3} are increasingly expected to support multiple abilities beyond next token generation. Besides producing answers, they often need to represent text as semantic vectors for retrieval~\citep{gao-etal-2021-simcse,behnamghader2024llmvec,zhang2025qwen3,feng-etal-2022-language,miao-etal-2024-enhancing,muennighoff-etal-2023-mteb} and compress long documents or interaction histories into compact states for context management and reduced computational and storage cost~\citep{chevalier-etal-2023-adapting,ge2024incontext,zhang2026agentic}. These abilities are important in many modern LLM applications, including retrieval-augmented generation (RAG)~\citep{NEURIPS2020_6b493230}, long context reasoning and agentic workflows. However, they are usually handled by separate models or separate modules. An embedding model produces retrieval vectors, a compressor shortens long contexts and a generator produces the final text. This separated design makes the system complicated and inefficient. Since these modules use different internal representations, their hidden states or KV cache cannot be directly reused and the same text may be processed multiple times across different stages.

The rise of reasoning language models~\citep{huang-chang-2023-towards,jaech2024openai,XU2025101370,muennighoff-etal-2025-s1,li-etal-2025-test,zeng-etal-2025-revisiting} makes this issue more important. Reasoning before answering has been shown to improve generation quality on complex tasks such as math and coding~\citep{lightman2024lets,miao2024improving,wei2022chain,miao2026neoamt,guo2025deepseek}. At the same time, reasoning can also help retrieval, since a query with explicit reasoning or decomposition may better capture the real information need~\citep{su2025bright,shao2025reasonir}. However, reasoning traces are often long, which increases the pressure on inference cost and makes compression necessary. These facts suggest that reasoning-driven generation, retrieval and compression are closely connected. 

Some recent works~\citep{muennighoff2025generative,10.1145/3774904.3792826,lan2026umer} try to unify part of this picture. For example, GritLM~\citep{muennighoff2025generative} trains one model for both generation and text embedding. 
However, it still uses different attention masks for the two modes, that is, bidirectional attention for embedding and causal attention for generation. This makes the two modes less natural to combine in one forward process. More importantly, these methods mainly focus on generation and embedding, while context compression and reusable latent memory are not directly studied. As a result, current systems still often need separate compressors, separate embedding models or expensive document level cache storage.



Based on the above observations, we explore how to unify three distinct yet related tasks, reasoning-driven generation, text embedding and context compression, in one forward pass for LLMs and propose a training framework named \methodname{}. 
Our model can serve as the retriever, generator and context compressor simultaneously in the RAG settings with the causal attention mask. This model greatly reduces the deployment effort and makes it possible to reuse the KV cache of different tasks for each other. It also makes a step towards a truly unified and reasoning-enhanced model leveraging the same internalized representations for three distinct tasks.
To support this process, we build a specialized KV cache server for storing and retrieving compressed document memories, and propose hybrid paged attention to manage two types of KV cache, regular prefix and dynamic KV cache, and compressed KV cache of \tokens{} in the constructed inference engine.

We highlight our main contributions as follows:
\begin{itemize}
	\item 
    First, we train one decoder only LLM to support text generation, embedding generation and context compression under the same causal attention mask. 
    This provides a simple way to unify three abilities that are usually trained and deployed separately.
	Our approach may shed light on the direction of a unified representation learning and inference paradigm for efficient latent memory-augmented generation and continual learning in which generation, semantic retrieval and context compression are conducted with the same internal representation of a single model. 
    
    \item Second, we introduces a new paradigm for text embeddings which contains mixed text and latent-based reasoning process, self-reason-latent-embed, where the model first generates text-based reasoning tokens, then switches to producing latent representations and finally generates the text representation by pooling the latent representations. This training framework also enables a new generation paradigm for RAG, latent memory-augmented generation, where the context is compressed, updatable latent memory/KV cache of \tokens{} rather than raw long document texts.
	\item We also propose \emph{{hybrid paged attention}} (HPA) to construct a new inference engine for our models. This new engine combines the idea of paged attention~\citep{10.1145/3600006.3613165} and our flexible inference paradigm enabled by our training framework. 
    This renders it a versatile LLM serving engine, capable of executing three tasks within a single forward pass while sustaining high throughput.
    As shown in Table~\ref{tab:detailed_actual_inference_time}, our new LLM inference engine achieves a 10× speedup over the baseline implementation on the same GPU.
	
\end{itemize}

%% file: sections/method.tex
\section{\methodname{}: unified representation learning for generation, retrieval and compression}
\label{sec:method}

\begin{figure}[ht]
	\centering
	\includegraphics[width=0.98\textwidth]{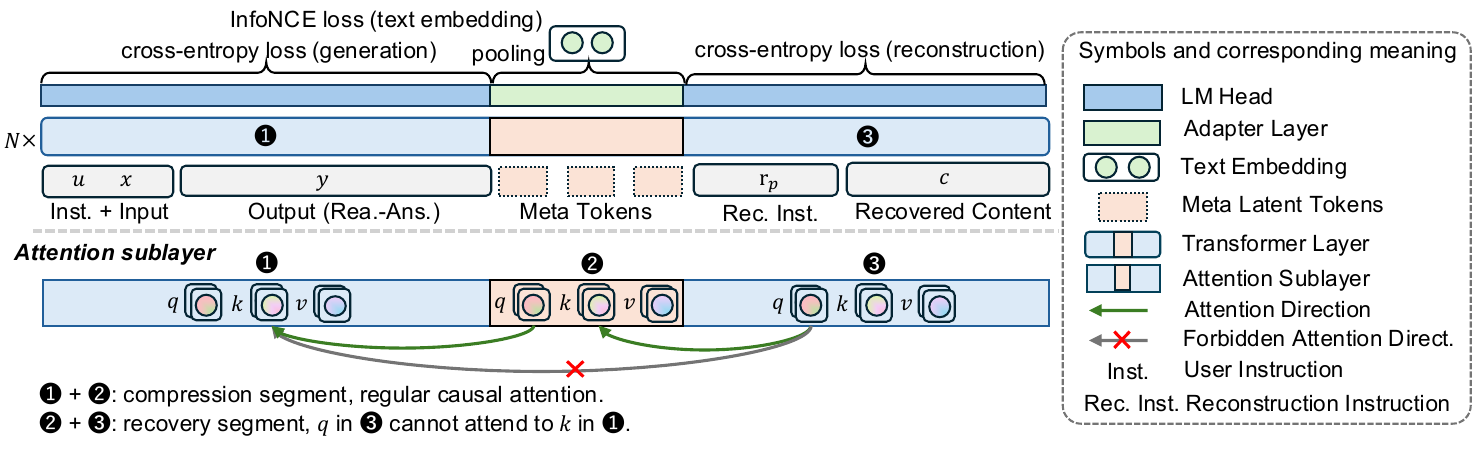}
    \caption{\methodname{} training framework. One forward pass of a single model can fulfill three objectives: (1) generating output (reasoning-answers); (2) compressing user-assistant chat history into compact representations/KV cache of \tokens{} through attentions; (3) obtaining reasoning-enhanced text embeddings of context by pooling the token representations of \tokens{}.  
    }
	\label{fig:grc_training}
\end{figure}



In this work, we aim to develop a model that unifies reasoning-driven generative, embedding and compression tasks in one forward pass with flexibility.
The training framework is shown in Figure~\ref{fig:grc_training}.

We use trainable \tokens{} $\{r_i\}_{i=1}^{m}, r_i \in \mathbb{R}^{d}$ to bridge three training tasks: text generation, context compression and text embedding where $d$ is the hidden dimension. $\{r_i\}_{i=1}^{m}$ are a small number of trainable parameters. 
Unlike gisting~\citep{mu2023learning} and other previous works~\citep{ge2024incontext}, 
the \tokens{} are non-intrusive for the language model (we do not insert \tokens{} into the model vocabulary). 
This design not only reduces the likelihood of models producing gibberish tokens in edge cases but also affords considerable flexibility at inference time.
This design further decouples the token embedding space, enabling each meta token to learn its own parameter weights during training.
As shown in Figure~\ref{fig:grc_training}, the representations of intermediate layers corresponding to \tokens{} are used for context compression and reconstruction.
The last hidden states corresponding to \tokens{} are further transformed by an adapter layer $A$ and a pooling operation to obtain text representation.
All these operations are conducted in a single forward pass.
These tokens play a role analogous to registers in computer architectures, caching small yet critical information to accelerate inference while exerting minimal impact on generative capability.

\paragraph{Training data.}
The original training dataset $\mathcal{D}$ consists of two main types of data: $\mathcal{D}_g$ and $\mathcal{D}_e$.
The first one $\mathcal{D}_g$ is reasoning-driven generative data, that is, user-assistant chat history with reasoning traces.
The training examples can be denoted as $(u, x, y)$ where $u$ and $x$ are the user instruction and query and $y$ is the model response including thinking and answer. 
The second one $\mathcal{D}_e$ is text retrieval data in which there are user instruction $u$, user query $x$ and a positive document $d_p$ related to the query and a list of negative documents $\{d_n^j\}$ where $1 \leq j \leq M-1$ and suppose we have $M$ documents for this $x$ in total.
For the second type of data, we use LLMs to generate reasoning traces $y$ for each query because we focus on the reasoning-driven paradigm. Thus, the user instruction $u$ and query $x$ are augmented into $(u, x, y)$.
To make each training example serve as three training signals, that is, generative, embedding and compression training signals, We make several adaptations to both types of training data.

\paragraph{Preparation for compression task.}
For the original generative data $\mathcal{D}_g$, the original user-assistant chat history $(u, x, y)$ serves as the first segment $\cnum{182}$ in Figure~\ref{fig:grc_training}. 
We append latent register tokens $\{r_i\}_{i=1}^{m}$ and a reconstruction instruction $r_p$ and recovered context $c$ into the sequence. $r_p$ and $c$ constitute the second segment $\cnum{184}$ in Figure~\ref{fig:grc_training}.  $r_p$ is randomly selected from a prompt set as shown in Table~\ref{app_table_recons_prompt_set}, such as ``What were we discussing earlier?''. Recovered context $c$ is the user instruction, query and model response $(u, x, y)$ in this case. Thus, the token sequence of a generative training example is $(u, x, y, \{r_i\}_{i=1}^{m}, r_p, c)$.

For the original embedding data $\mathcal{D}_e$, we also append \tokens{} $\{r_i\}_{i=1}^{m}$ and a reconstruction instruction $r_p$ and recovered context $c$ after reasoning enhanced queries $(u, x, y)$, positive document $d_p$ and a list of negative document $\{d_n^j\}$.
The token sequences of a query, positive document and negative documents are $(u, x, y, \{r_i\}_{i=1}^{m},  r_p, c)$, $(u, d_p, y, \{r_i\}_{i=1}^{m},  r_p, c)$, $(u, d_n, y, \{r_i\}_{i=1}^{m},  r_p, c)$ respectively in which $u$ in document instances is a different user instruction, such as ``Represent this text'' and $y$ is usually ``None''. The contexts $c$ for queries and positive and negative documents are $(u, x, y)$, $d_p$, and $d_n$ respectively. 

\paragraph{Preparation for retrieval task.}
Now that, we have augmented the training data $\mathcal{D}_g$ and $\mathcal{D}_e$ for the compression task.
To reuse $\mathcal{D}_g$ for text retrieval training, we need to prepare positive and negative documents for each generative training example $(u, x, y, \{r_i\}_{i=1}^{m}, r_p, c)$.
Following the unsupervised sentence embedding~\citep{gao-etal-2021-simcse}, we use the training example itself $(u, x, y, \{r_i\}_{i=1}^{m}, r_p, c)$ as the positive document. The other in-batch training examples that could be a augmented generative training example or embedding training example, are utilized as negative documents.

Through the above preparations, we have unified the two types of data into a unified format: each training example $x$ have one query instance $q$, and a positive instance $d_p$ and negative instances $\{d_n\}$.
For unified generation training data, the training example itself is used as the query and positive instances, the negative instances are randomly selected from the mini-batch of the training dataset during training.
The embedding training data requires no further modification.

Given an hidden state sequence $(\bm{u}_1, \ldots, \bm{x}_w, \ldots, \bm{y}_k)$ and latent register tokens $\{r_i\}_{i=1}^{m}$ where $\bm{u}$, $\bm{x}$ and $\bm{y}$ are user instruction, user input and assistant response token representations, 
the input hidden state sequence of our model is:
\begin{equation}
\label{eq:input}
    (\underbrace{\bm{u}_1, \ldots, \bm{u}_w, \bm{x}_{w+1}, \ldots,  \bm{x}_{j}, \bm{y}_{j+1}, \ldots, \bm{y}_k}_{\textrm{first segment: \cnum{182}}}, \underbrace{\bm{r}_{k+1}, \dots, \bm{r}_{k+m}}_{\textrm{latent register tokens: \cnum{183}}}, \underbrace{\bm{q}_{k+m+1}, \ldots, \bm{q}_{p}, \bm{c}_{p+1}, \ldots, \bm{c}_n}_\textrm{second segment: \cnum{184}}), 
\end{equation}
where $\bm{q}$, $\bm{c}$ are reconstruction instruction and ideal recovered context token representations. 

For generative training, we use the cross-entropy loss on the output hidden states of the first segment except the \tokens{} $\{r_i\}_{i=1}^{m}$, which can be expressed as follows:
\begin{align}
    \mathcal{L}_{\text{Gen}} = -\frac{1}{k-j}\sum_{i=j+1}^{k} \log P(f_{\theta, \eta}(y^{(i)}) | f_{\theta, \eta}({u}, {x}, {y}^{(<i)})),
\end{align}
where $f_{\theta, \eta}$ is the \methodname{} model with model parameters $\theta$ and language head $\eta$. 
Note that $\mathcal{L}_{\text{Gen}}$ is also applied to positive and negative documents, where $u$ is the user instruction (e.g., ``Represent this text: \{doc\}''), $x$ is the document, and $y$ is the model response, which is set to ``None''.
For compression and reconstruction tasks, we mask out the $k$ vectors in the first segment \cnum{182} when computing attention scores from the segment \cnum{184} so that $q$ vectors in segment \cnum{184}  only attend to the $k$ vectors of \tokens{} $\{r_i\}_{i=1}^{m}$ in segment \cnum{183} while $q$ vectors of \tokens{} can attend to the $k$ vectors in segment \cnum{182}:

\begin{align}
    \mathcal{L}_{\text{Recons}} = -\frac{1}{n-p}\sum_{i=p+1}^{n} \log P(f_{\theta, \eta}(c^{(i)}) | f_{\theta, \eta}(\text{mask}(\text{segment \cnum{182}}), \{r\}, {q}, {c}^{(<i)})).
\end{align}
Note that $\mathcal{L}_{\text{Recons}}$ is also trained for positive and negative documents in which $c$ denotes the documents.
By masking out the token representations in segment \cnum{182} in attention computation, the model need to learn how to reconstruct the segment \cnum{182} by probing $k$ and $v$  representations of \tokens{} in segment \cnum{183}. Through this process, we compress the semantic information in segment \cnum{182} into the $k$ and $v$ representations of \tokens{}. Note that the $\mathcal{L}_{\text{Gen}}$ and $\mathcal{L}_{\text{Recons}}$ can be computed in one forward and backward pass with a customized causal attention mask simultaneously, unlike previous studies~\citep{ge2024incontext} using LLMs as the encoder and decoder separately.

For embedding training, we extract the last hidden states $h_{r_i}, i \in [k+1, k+m]$ of \tokens{} $\{r_i\}$ and apply an adapter layer $A$ to them: ${a}_{r_i} = A(h_{r_i})$. Then we use the mean pooling and normalization operation to obtain the final text embedding: $e = \text{norm}(\text{pooling}({a}_{r_i}))$.
Unlike GritLM that trains embedding and generative tasks with two separate training datasets, we also utilize the training data for generative tasks for embedding training.
Thus, we have a unified training data example that consists of query $q$, positive instance $d_p$ and negative instances $\{d_n\}$ for both generative and embedding training data. We can apply contrastive learning~\citep{pmlr-v119-chen20j,gao-etal-2021-simcse} on the text embedding representations:
\begin{align}
    \mathcal{L}_{\text{Rep}} = -\frac{1}{M}\sum_{i=1}^M\log\frac{\exp(\tau \cdot \sigma(e_q^{i}, e_{d_p}^{i}))}{\sum_{j=1}^M\exp(\tau \cdot \sigma(e_q^{i}, e_d^{j}))},
\end{align}
where $d \in \{d_p, d_n\}$ and $\tau$ represents a temperature hyperparameter. $\sigma$ corresponds to the cosine similarity operation.
We extract the hidden states $h_{r_i}$ of \tokens{} that already have the compressed semantic information of the segment \cnum{182} which is obtained by the masking operation and reconstruction loss $L_\text{recons}$. 
We then apply an adapter layer $A$ with a pooling operation on the $h_{r_i}$ to obtain the final text embedding $e$. In this way, we preserve the semantic information of the first segment through $\mathcal{L}_{\text{Recons}}$ and further transform it into the final text embedding through $\mathcal{L}_{\text{Rep}}$, which achieves almost perfect compatibility between context compression and contrastive representation learning. Another benefit is that the token representations for generative tasks are not affected by the embedding training anymore.
The training loss will be:
\begin{align}
    \mathcal{L} = \alpha \cdot \mathcal{L}_{\text{Gen}} + \beta \cdot \mathcal{L}_{\text{Recons}} + \gamma \cdot \mathcal{L}_{\text{Rep}}.
\end{align}

%% file: sections/inference.tex
\section{Flexible inference}
\label{sec:inference}

\begin{wrapfigure}{r}{0.42\textwidth}
	\centering
	\includegraphics[width=0.4\textwidth]{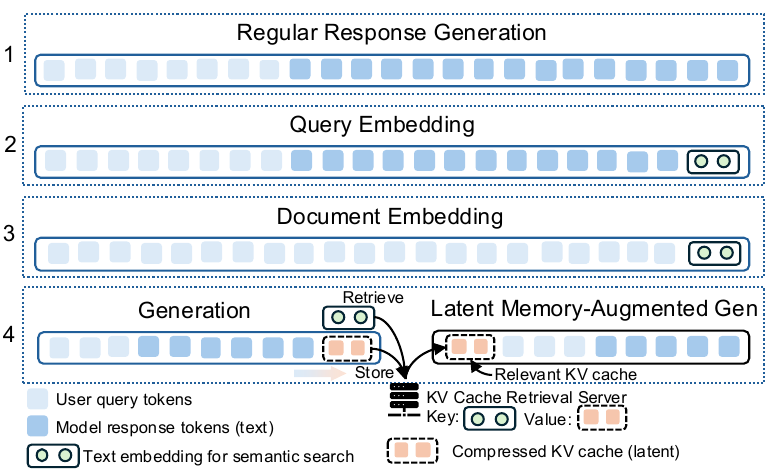}
	\caption{Four diverse generation patterns: (1) regular generation; (2) self-reason-latent-based query embedding; (3) document embedding and (4) latent memory-augmented generation.
    }
	\label{fig:grc_inference_patterns}
\end{wrapfigure}

A single \methodname{} model can enable flexible inference across four generation patterns as shown in Figure~\ref{fig:grc_inference_patterns}. Three tasks (generation, embedding and compression) are seamlessly conducted into one forward pass in a flexible way with the same causal attention mask via \tokens{}. The naive inference implementation of \methodname{} models are described in Appendix~\ref{app:naive}.
\paragraph{Difference with GritLM.} 
The actual inference implementation of GritLM loads AutoModel version for embedding tasks and AutoModelForCusalLM for generative tasks\footnote{\url{https://github.com/ContextualAI/gritlm/blob/971068105a8508bca421841c59fddba7f6596402/gritlm/gritlm.py\#L24}}, which means that we need to host two model replica in device memory if we want to obtain the response and text embedding for a user query.
Though this problem might be mitigated by carefully modifying the code, GritLM can only generate either a text response or vector for embedding in one forward pass because of the attention difference of GritLM between the generation and embedding modes (bidirectional for text embedding and unidirectional for text generation). 
Our models use the causal mask for both cases and can finish three tasks in one forward pass at any position in the sequence. 

\begin{wrapfigure}{r}{0.3\textwidth}
    \centering
    \includegraphics[width=0.29\textwidth]{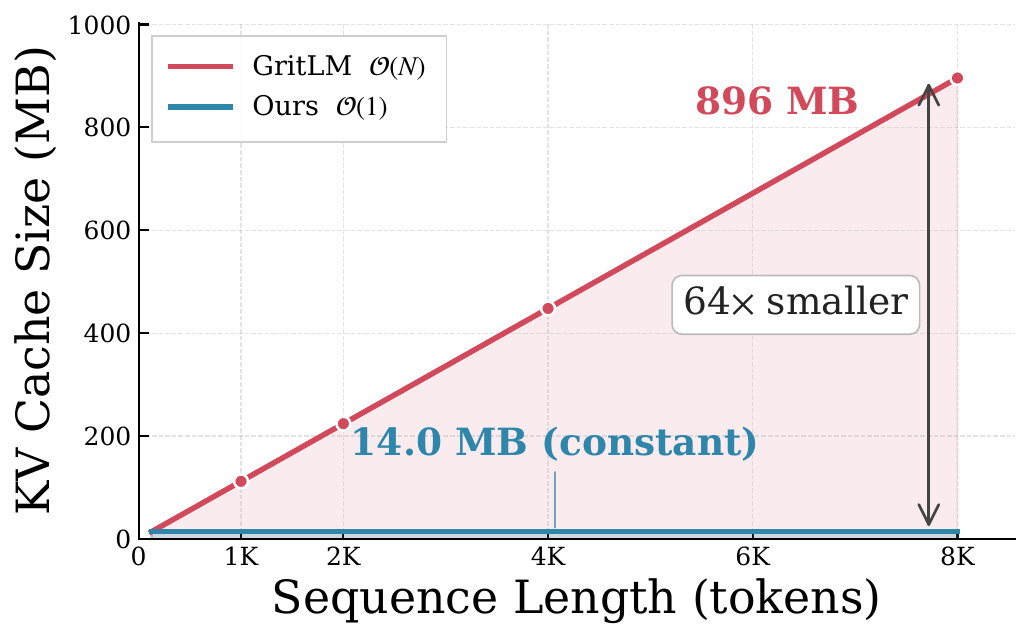}
    \caption{Comparison of document KV cache storage size.}
    \label{fig:kv_cache_size_comparison}
\end{wrapfigure}

\begin{wrapfigure}{r}{0.44\textwidth}
	\centering
	\includegraphics[width=0.42\textwidth]{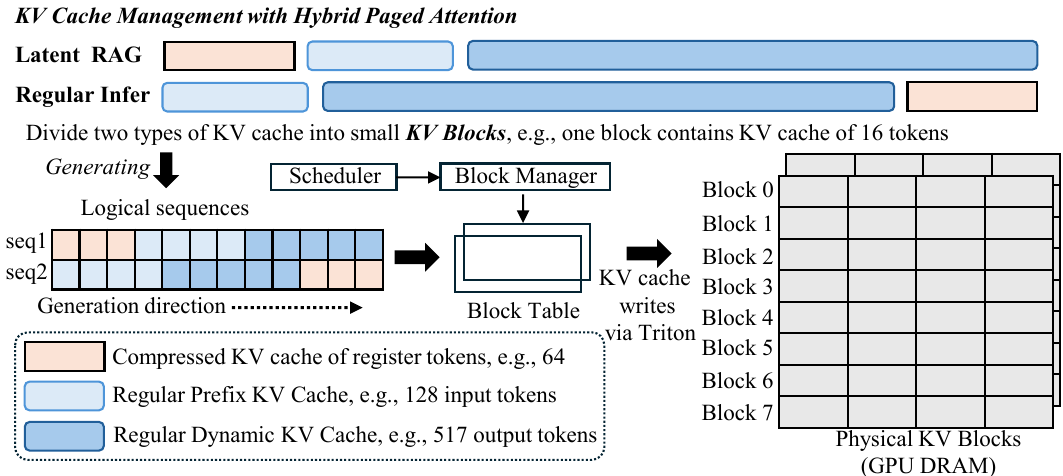}
	\caption{KV cache management with the proposed \textbf{hybrid paged attention} for speeding up inference.}
	\label{fig:hpa}
\end{wrapfigure}

\paragraph{KV cache cost.}
One advantage of our method is the reduced storage cost of document KV cache.
The KV cache size for GritLM is computed as:
\begin{equation}
\mathrm{KV\ size} = 2 \times L \times H_{kv} \times d_{h} \times N \times \mathrm{bytes},
\end{equation}
where $L$ denotes the number of transformer layers, $H_{kv}$ the number of key-value heads, $d_{h}$ the head dimension, $N$ the sequence length, and $\mathrm{bytes}$ the number of bytes per element (e.g., $2$ for bfloat16). The factor of $2$ accounts for storing both keys and values.
Suppose we use the Qwen3-1.7B model architecture, that is, $L$ is $28$, $H_{kv}$ is $8$, $d_h$ is $128$, and $\mathrm{bytes}$ is 2 (bfloat16),
thus the relationship of KV cache size $\mathrm{Y}$ (MiB) with the sequence length $N$ is $\frac{114688N}{1024^2}$.
The document is compressed into the KV cache of latent register tokens in our method, the number of latent register tokens is $N_r$ which is a fixed number, for example, $128$.
The KV cache size will be $\frac{114688 \times N_r}{1024^2}$.
This KV cache size also reduces the computational cost.
The comparison of the KV cache storage cost between \methodname{} and \grit{} under different document lengths is shown in Figure~\ref{fig:kv_cache_size_comparison} where the number of \tokens is 128.
\paragraph{Hybrid paged attention for LLM serving.}
To further speed up the inference, we propose hybrid paged attention (HPA) to construct a new inference engine.
This proposed method is based on paged attention~\citep{10.1145/3600006.3613165}. Paged attention is an attention algorithm inspired by the virtual memory and paging techniques in operating systems for LLM serving. It achieves high throughput serving by pre-allocating KV cache in the device memory and partitioning them into fixed-size non-continuous blocks.
In our model's inference, the context can be compressed into compressed KV cache with $O(1)$ length. Thus, we have two types of KV cache. One is regular KV cache including prefix KV cache for user prompts and regular dynamic KV cache for model responses. The other is the compressed KV cache for the context as shown in Figure~\ref{fig:hpa}. In our HPA approach, we not only put the prefix and regular dynamic KV cache into the blocks, but also store the KV cache of \tokens{}, that is, the compressed KV cache, into the corresponding blocks via Triton operations. This approach allows us to retain the benefits of paged blocks during model serving and our models are still capable of carrying out the three tasks in one forward pass on our new inference engine.

%% file: sections/exp.tex
\section{Experiments}
\label{sec:exp}

\subsection{Experimental setting}

\paragraph{Training data.}
Various reasoning-based question-answering (QA) and retrieval data are utilized for the model training. The training dataset consists of general reasoning-based QA pairs, reasoning-intensive queries with positive and negative documents, minor agentic data. We collected approximately 600K training examples, of which only around 20\% were used during training.
More details can be found in Appendix~\ref{app:training_data}.

\paragraph{Models.}
Qwen3-1.7B-Base and Qwen3-4B-Base~\citep{yang2025qwen3} are utilized as base models for training. More training details can be found in Appendix~\ref{app:train_details}.

\paragraph{Evaluation.} 
We evaluate our models across the following task categories: for retrieval, we use the BRIGHT benchmark~\citep{su2025bright}; for generation, the GSM8K~\citep{cobbe2021training} and BBH~\citep{suzgun-etal-2023-challenging} benchmarks; and for document compression, the PwC dataset~\citep{ge2024incontext} together with a newly curated set of Wikipedia-based markdown documents. The latter were constructed from articles dated between January 1 and March 1, 2026 to minimize the risk of data contamination.
We also test our models on the RAG setting and our new generation paradigm, latent memory-augmented generation with the Natural Question (NQ) dataset~\citep{kwiatkowski-etal-2019-natural} and its BEIR NQ corpus~\citep{thakur2021beir}.
Furthermore, we conduct latency evaluation for comparison between the naive inference implementation and our HPA-based inference engine.

\subsection{Experimental results}
\subsubsection{Reasoning-intensive retrieval tasks}
\begin{table}[ht]
	\caption{Retrieval performance on the BRIGHT benchmark. No external tools/modules are used. The scores of nDCG@10 metric are reported for all datasets:
		Biology (Bio.), Earth Science (Earth.), Economics (Econ.), Psychology (Psy.), Robotics (Rob.),
		Stack Overflow (Stack.), Sustainable Living (Sus.), LeetCode (Leet.), Pony, AoPS, TheoremQA with
		question retrieval (TheoQ.) and with theorem retrieval (TheoT.).
		Models are introduced in Table~\ref{tab:models}. \methodname{} query prompt is available in Table~\ref{tab:query_prompt}.
	}
	
	\label{tab:bright_result}
	\centering
	\resizebox{0.95\linewidth}{!}{%
		\begin{tabular}{l*{14}{c}}
			\toprule
			\multirow{2.5}{*}{\textbf{Model}} & \multirow{2.5}{*}{\textbf{Size}} & \multicolumn{7}{c}{\textbf{StackExchange}}  & \multicolumn{2}{c}{\textbf{Coding}} & \multicolumn{3}{c}{\textbf{Theorem-based}} & \multirow{2.5}{*}{\textbf{Avg.}} \\
			\cmidrule(lr){3-9} \cmidrule(lr){10-11} \cmidrule(lr){12-14}
			
			& & \textbf{Bio.} & \textbf{Earth.} & \textbf{Eco.} &  \textbf{Psy.}  & \textbf{Rob.} & \textbf{Stack.} & \textbf{Sus.} & \textbf{Leet.} & \textbf{Pony}  & \textbf{AoPS} & \textbf{TheoQ.} & \textbf{TheoT.} & \\ 
			\midrule
			{BM25}  		                    & N/A  & 18.9 & 27.2 & 14.9 & 12.5 & 13.6 & 18.4 & 15.0  & 24.4 & 7.9  & 6.2 & 10.4 & 4.9 & 14.5 \\
			\midrule
			\multicolumn{15}{c}{\textit{Proprietary models}} \\
            \midrule
            {OpenAI}  		                    & N/A & 23.3 & 26.7 & 19.5 & 27.6 & 12.8 & 14.3 & 20.5  & 23.6 & 2.4  & 8.5 & 23.5 & 11.7 & 17.9 \\
			{Cohere} 		                    & N/A & 18.7 & 28.4 & 20.4 & 21.6 & 16.3 & 18.3 & 17.6 & 26.8 & 1.9 & 6.3 & 15.7 & 7.2 & 16.6 \\
			{Google} 		                    & N/A & 22.7 & 34.8 & 19.6 & 27.8 & 15.7 & 20.1 & 17.1 & 29.6 & 3.6 & 9.3 & 23.8 & 15.9 & 20.0\\
            {Voyage}                            & N/A & 23.1 & 25.4 & 19.9 & 24.9 & 10.8 & 16.8 & 15.4 & 30.6 & 1.5 & 7.5 & 27.4 & 11.6 & 17.9 \\
			\midrule
            \multicolumn{15}{c}{\textit{Open-sourced models}} \\
            \midrule
            Inst-XL                            & 1.5B & 21.6 & 34.3 & 22.4 & 27.4 & 18.2 & 21.2 & 19.1 & 27.5 & 5.0 & 8.5 & 15.6 & 5.9  & 18.9 \\
            E5                                 & 7.1B & 18.6 & 26.0 & 15.5 & 15.8 & 16.3 & 11.2 & 18.1 & 28.7 & 4.9 & 7.1 & 26.1 & 26.8 & 17.9 \\
            SFR                                & 7.1B & 19.1 & 26.7 & 17.8 & 19.0 & 16.3 & 14.4 & 19.2 & 27.4 & 2.0 & 7.4 &  24.3 & 26.0 & 18.3 \\
            Qwen                               & 7.7B & 30.6 & 36.4 & 17.8 & 24.6 & 13.2 & 22.2 & 14.8 & 25.5 & 9.9 &  14.4 & 27.8 & 32.9 & 22.5 \\
            {ReasonIR-8B}                      & 8B & 26.2 & 31.4 & 23.3 & 30.3 & 18.0 & {23.9} & 20.5  & 35.0 & 10.5  & {14.7} & 31.9 & {27.2} & {24.4} \\
			{\grit{}-7B}                       & 7.1B & 24.8 & 32.3 & 18.9 & 19.8 & 17.1 & 13.6 & 17.8 & {29.9} & {22.0} & 8.8 & 25.2 & 21.2 & 21.0 \\
			\midrule
            {\methodname{}-1.7B (T0.6, 4K)}       & 1.7B & {37.5} & {37.6} & {24.5} & {31.3} & {18.9} & {23.1} & {25.7}  & 26.2 & {11.2}  & {9.0} & {37.3} & {26.0} & {25.7}  \\ 

            
            \rowcolor[rgb]{0.925,0.957,1} {\methodname{}-4B (T0.6, 4K)} & 4B  & 33.5 & 32.0 & {26.4} & 30.6 & {19.6} & {28.2} & 24.8 & {38.3} & 4.1 & 11.3 & 44.2 &  {33.9} & \textbf{27.2} \\
			
			\bottomrule
		\end{tabular}%
	}
	
\end{table}
BRIGHT~\citep{su2025bright} is a challenging text retrieval benchmark that requires intensive reasoning to retrieve relevant documents. 
We report nDCG@10 scores as the main metric.
For user queries, we use the query prompt in Table~\ref{tab:query_prompt} to encourage the model to generate reasoning tokens first. We then use the mean pooling operation on the representations of \tokens{} to obtain the embedding.
For documents, we use the document prompt in Table~\ref{tab:doc_prompt} and directly obtain the embedding via the mean pooling operation on the representations of \tokens{}.
Table~\ref{tab:bright_result} demonstrates that even our small model (1.7B) can beat ReasonIR-8B~\citep{shao2025reasonir}, demonstrating the effective of our new embedding generation paradigm.
\begin{figure}[h]
	\centering
	\includegraphics[width=0.98\textwidth]{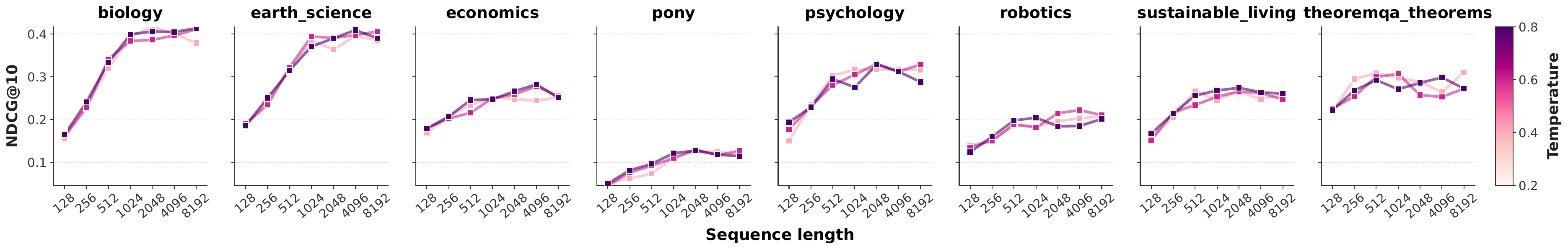}
	\caption{NDCG@10 scores of \methodname{}-1.7B with different max new tokens (128, 256, 512, 1024, 2048, 4096 and 8192) and varying temperatures (0.2, 0.4, 0.6, 0.8) on BRIGHT retrieval tasks.}
    \label{fig:emb_ablation_study}
\end{figure}
We further investigate the impact of reasoning lengths and sampling temperatures during reasoning on the performance as shown in Figure~\ref{fig:emb_ablation_study}.
The model's score increases as the number of generated tokens grows, especially for the biology, psychology and earth science tasks. However, once generated tokens reach a certain length, the performance plateaus as shown in Figure~\ref{fig:emb_ablation_study}.
This indicates that self-generated reasoning traces are beneficial for the retrieval performance but overthinking does not contribute much.
We also find that the performance is better if we do not add <eos> token before the pooling operation to obtain the final text embedding.






\subsubsection{Generative tasks}
We utilize BIG-Bench Hard (BBH)~\citep{srivastava2023beyond,suzgun-etal-2023-challenging} and GSM8K~\citep{cobbe2021training} for evaluating the generative peroformance of models. BBH is a diverse evaluation benchmark based on BIG-Bench for evaluating the general reasoning capabilities of LLMs, which consists of 23 challenging tasks.
GSM8K contains a set of math problems that require reasoning to solve.
As shown in Table~\ref{tab:gen_result}, our models maintains competitive performance on generative tasks requiring reasoning.


\subsubsection{Document compression.}
\begin{figure}
    \centering
    \includegraphics[width=0.98\linewidth]{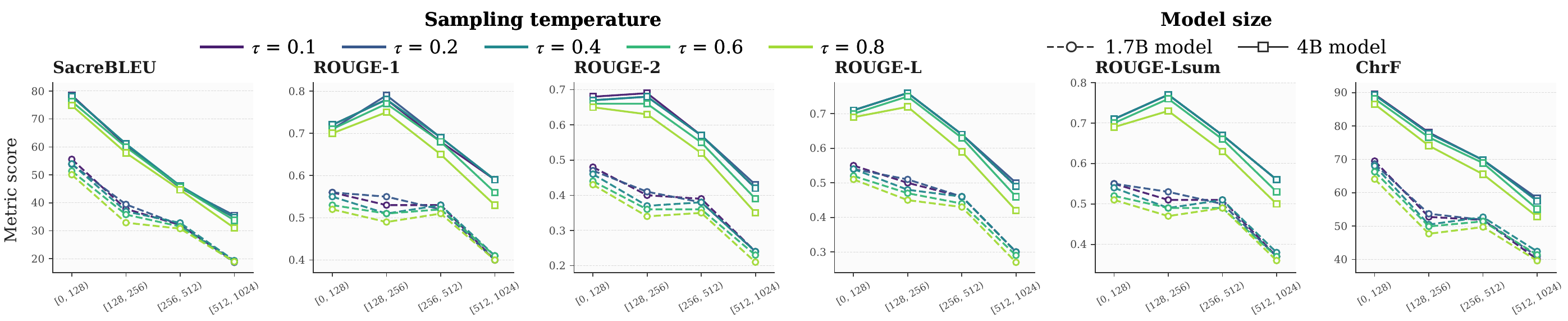}
    \caption{Reconstruction scores of \methodname{} models on the wikipedia markdown documents compression task. The X-axis denotes the document length ranges.}
    \label{fig:wiki_recons_scores}
\end{figure}

We use the test set of the PwC\footnote{\url{https://huggingface.co/datasets/sggetao/PwC}} dataset~\citep{ge2024incontext} for document compression. The PwC dataset consists of (context, prompt, responses) triples, built for training and testing the context compression performance of models. Note that \methodname{} models do not use the training set of the PwC dataset in the training stage. Thus this evaluation can be considered as the out-of-domain testing. We report various metrics including sacrebleu, rouge and chrF. Table~\ref{tab:pwc_document_compression_reconstruction} presents the reconstruction results on the documents of the PwC test set of different models.
The other document compression evaluation task is based on the Wikipedia documents. 
\begin{table}[h]
	\caption{The results of PwC document compression and reconstruction task.}
	\label{tab:pwc_document_compression_reconstruction}
	\small
	\centering
    \resizebox{0.8\linewidth}{!}{%
	\begin{tabular}{{l}*{6}{c}}
		\toprule
		& \textbf{sacrebleu} & \textbf{rouge1} & \textbf{rouge2} & \textbf{rougeL} & \textbf{rougeLsum} & \textbf{chrF}  \\
		\midrule
        AutoCompressor-7B~\citep{chevalier-etal-2023-adapting} &  0.84 & 0.09  & 0.01  & 0.07 & 0.09 & 17.72 \\
		ICAE-v2-7B~\citep{ge2024incontext} & 17.98 & 0.55 &  0.33 & 0.39 & 0.50 & 44.67   \\
		
		\methodname{}-1.7B & 18.34 & 0.41   &0.25  &  0.31 & 0.38 & 47.37  \\
		
		\rowcolor[rgb]{0.925,0.957,1} \methodname{}-4B & \textbf{38.06} & \textbf{0.63} & \textbf{0.48} & \textbf{0.53} & \textbf{0.60} & \textbf{66.63}   \\
		
		\bottomrule
	\end{tabular}}
\end{table}
Figure~\ref{fig:wiki_recons_scores} shows the reconstruction scores of the Wikipedia document compression task under different sampling temperatures, model sizes and document lengths. 
Our findings are as follows: (1) the 4B model significantly outperforms the 1.7B model across all metrics and sequence lengths, with the performance gap being most pronounced in the short-length regime; (2) as sequence length increases, all metrics decline; however, SacreBLEU and ChrF exhibit the steepest decreases; (3) the effect of temperature is relatively modest; $\tau = 0.8$ shows the most pronounced degradation in the long-context regime, as expected.

		
		
		

\subsubsection{RAG}

Following previous studies~\citep{muennighoff2025generative}, we use Natural Questions dataset~\citep{kwiatkowski-etal-2019-natural} with the splitting method\footnote{\url{https://github.com/ContextualAI/gritlm/blob/main/rag/prepare_qa.py}} of \grit{} and randomly select 500 examples from the test split for the following evaluation. The 2,681,468 documents from BEIR NQ corpus~\citep{thakur2021beir} are utilized as the retrieval source.
For evaluation metrics, we use the ``match'' metric\footnote{\url{https://github.com/ContextualAI/gritlm/blob/main/rag/tasks/evaluation.py}}. 
For \methodname{}, temperature is set to $0.1$ in all cases. When computing query embeddings, the max new tokens is set to 4096 while the max new tokens is set to 16 for computing document embeddings.
The max new tokens is set to 1024 when the model generates answers given the retrieved documents for NQ dataset.
The result is shown in Table~\ref{tab:rag_results}. ``No doc'' denotes standard response generation without retrieval. ``w/. compressed doc'' indicates that the context is provided as a compressed KV cache of \tokens{}. ``w/. plain text doc'' corresponds to the standard RAG setting.
We find that the result of RAG setting with regular retrieved document with \methodname{}-4B model is better than \grit{}-7B though we do not finetune our models on the embedding training data like NQ or E5~\citep{wang2024multilingual}. This demonstrates the generalization ability of our training method.
The result in Table~\ref{tab:rag_results} also indicates that the latent memory, that is, the compressed KV cache indeed carries document information and can be identified and utilized by our models.

\begin{table}[t]
  \centering
  \begin{minipage}{0.6\textwidth}
    \centering
    \caption{Comparison of different methods on the RAG setting using the Natural Question (NQ) dataset. 
We suppose the document sequence length is $N$.
Best results are highlighted in \textbf{bold}.}
\label{tab:rag_results}
\resizebox{0.88\linewidth}{!}{
    \begin{tabular}{l c c}
        \toprule
        {\textbf{Method}} & \textbf{NQ (\%)} & \textbf{Doc KV cache size} \\
        \midrule
        GritLM-7B (no doc)                       & 21.40  & 0 \\
        GritLM-7B                                & 31.60  & $O(N)$ \\
        \midrule
        
        \methodname{}-1.7B (no doc)             & 26.20   & 0 \\
        \methodname{}-1.7B (w/. compressed doc)   & 29.00  & $O(1)$ \\
        \methodname{}-1.7B (w/. plain text doc)   & 33.60   & $O(N)$ \\
        \midrule
        \methodname{}-4B (no doc)             & 35.40     & 0 \\
        \methodname{}-4B (w/. compressed doc)     & 40.40   & $O(1)$ \\
        \rowcolor[rgb]{0.925,0.957,1} \methodname{}-4B (w/. plain text doc)     & \textbf{45.80}  & $O(N)$ \\
    
        \bottomrule
    \end{tabular}}
  \end{minipage}
  \hfill
  \begin{minipage}{0.35\textwidth}
    \centering
    \caption{Performance on generative tasks.}
    \label{tab:gen_result}
    \resizebox{0.98\linewidth}{!}{
    \begin{tabular}{lcc}
    \toprule
       \textbf{Model}  & \textbf{GSM8K} & \textbf{BBH} \\ \midrule
        GPT-4-0613 & 95.0 & 89.1 \\
        GPT-J 6B & 2.5 & 30.2 \\
        Llama 2 Chat 70B & 59.0 & 49.0 \\
        Zephyr 7B $\beta$ & 28.0 & 44.9 \\
        Mistral 7B & 44.5 & 55.6 \\
        Mixtral 8x7B Instruct & 65.0 & 55.9 \\
       \midrule
       \grit{}-7B & 57.5 & 54.8 \\
       \midrule
       \methodname{}-1.7B & 61.79 & 43.03\\
       \rowcolor[rgb]{0.925,0.957,1} \methodname{}-4B & \textbf{67.40} & \textbf{64.27} \\
    \bottomrule
    \end{tabular}}
  \end{minipage}
\end{table}

\subsubsection{Hybrid paged attention performance}

\begin{figure}
    \centering
    \includegraphics[width=0.9\linewidth]{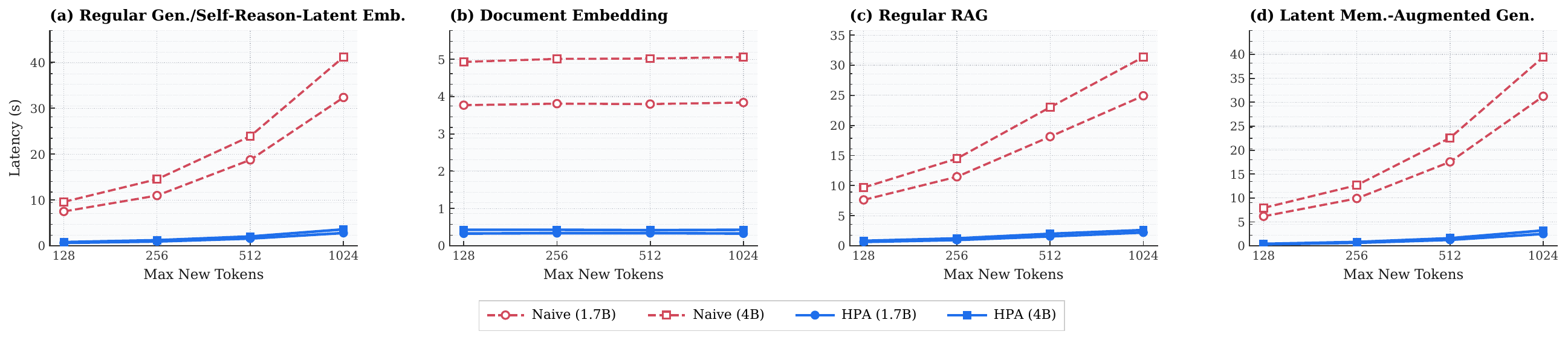}
    \caption{Comparison of latency (average consumed time (s) for one user query). Note that the latency of regular generation pattern is the same as the one for self-reason-latent embedding (reasoning-enhanced query embedding) because we use the same user queries and inference implementation for both patterns. Our implementation can return both generated response and query embeddings simultaneously. Document embedding can be considered as a non-reasoning-enhanced embedding. This test is conducted on one NVIDIA A100 GPU with 80GB device memory. The batch size is set to 1 for naive inference implementation. The hyperparameters used in the HPA-based LLM servering system is presented in Table~\ref{tab:hpa_hyperparameters}, Appendix~\ref{app:hpa_hyperparameters}. Detailed actual inference time can be found in Table~\ref{tab:detailed_actual_inference_time}.}
    \label{fig:hpa_result}
\end{figure}

Figure~\ref{fig:hpa_result} presents the latency evaluation result with the naive and HPA inference on different generation patterns with varying maximum generation lengths.
The hybrid paged attention significantly speeds up the inference speed of our models, especially when the max new tokens is set to a large number.
The difference between 1.7B and 4B models when using HPA is not so large. The detailed actual inference time can be found in Table~\ref{tab:detailed_actual_inference_time}, Appendix~\ref{app:eval_details}.



%% file: sections/related.tex
\section{Related work}
\label{sec:related_work}

\paragraph{Unifying generation and embedding.}
Text retrieval~\citep{10.1145/3637870,karpukhin-etal-2020-dense} and generation are inherently complementary, which has motivated substantial research effort into improving their interplay~\citep{pmlr-v119-guu20a}, RAG being a prominent example~\citep{NEURIPS2020_6b493230}, where information retrieved from external sources is leveraged to supplement or correct potential errors in the text generation process. However, these two tasks are typically handled by two distinct LLMs, trained with different objectives: retrieval models commonly employ bidirectional attention with contrastive learning~\citep{gao-etal-2021-simcse,pmlr-v119-chen20j,behnamghader2024llmvec}, whereas generative models rely on causal attention and cross-entropy loss. Moreover, the training data for the two tasks generally differ as well. GritLM~\citep{muennighoff2025generative} makes a direct attempt to unify these two disparate tasks within a single LLM by using different attention masks and separately curated training data for each, achieving promising results. 
However, it also suffers from the following problems: (1) \emph{sub-optimal training data utilization}: the generative data and embedding are prepared and trained separately. 
	 (2) \emph{heterogeneous attention mechanisms} for generative and embedding tasks. This causes the one forward pass can only finish one type of tasks, either generative tasks or embedding tasks. 
	 (3) \emph{prohibitive $O(N)$ storage cost of KV cache for document caching in RAG};
	 (4) \emph{reasoning} is not explored in the unified training case.
There are also some other works focusing on combining reasoning and embedding~\citep{10.1145/3774904.3792826,lan2026umer}. However, these studies usually omit the context compression perspective, especially when the reasoning traces are long.


\paragraph{Context compression.}
Context compression~\citep{li-etal-2023-compressing,li-etal-2025-prompt} reduces inference cost and is often essential for managing context in agentic and reasoning tasks~\citep{zhang2026agentic}. Existing approaches~\citep{li-etal-2023-compressing,chevalier-etal-2023-adapting, ge2024incontext} can be broadly categorized into text-level compression~\citep{jiang-etal-2023-llmlingua,pan-etal-2024-llmlingua} and latent-space compression~\citep{chevalier-etal-2023-adapting,ge2024incontext}. Text-level compression is typically achieved through prompting, whereas 
latent-based context compression methods usually consists of an encoder and a decoder, in which, the encoder converts context into the last hidden states of memory tokens and the decoder receives the last hidden state and continually generates text. This line of studies include using another language model (encoder-based or decoder-based)~\citep{NEURIPS2024_c5cf13bf,liu2025context}, image encoders~\citep{wei2025deepseek} to compress context. Our work focuses more on using attention to compress context and how to reuse the KV cache in one forwad pass for the three tasks that we are unifying, which is more aligned with gisting~\citep{NEURIPS2023_3d77c6dc}. 

%% file: sections/conclusion.tex
\section{Conclusion}
\label{sec:conclusion}

In this paper, we explore the possibility of unifying three objectives in one forward pass so that reasoning-driven generation, semantic retrieval and context compression can be conducted by the unified representation from a single model.
We use \tokens{} to decouple the dual-roles of regular tokens in previous studies.
The intermediate representations of \tokens{} are utilized to store the compressed semantic information of the context. The representation from the top Transformer block of \tokens{} is leveraged for extracting text embedding.
Furthermore, we use the same causal attention mask for all three tasks except that we mask out segment \cnum{182} when computing attention weights for segment \cnum{184}.
This design enables the flexible LEGO-style inference.
Extensive experiments spanning reasoning-intensive retrieval, generative tasks, document compression, latency analysis, and RAG settings validate the effectiveness of our approach.
We believe our approach takes a meaningful step toward the long-standing vision of a unified model capable of addressing nearly all NLP tasks, a direction we will continue to pursue.

\paragraph{Limitations.}

In our current implementation, we extract the last hidden states of \tokens{} from the final Transformer block and apply an adapter layer and a pooling operation to obtain the final text embedding. There might be a conflict that may affect the quality of text embeddings when we use the last token pooling. The reason is that, in this case, the hidden state of last meta latent token is also used to prediction the next token, which is usually a BOS token.
These two objectives may affect each other and introduce instability.


%% file: sections/author_contribution.tex
\section*{Author Contributions}
Author contributions are described below using the CRediT taxonomy.

\textbf{Zhongtao Miao:} Conceptualization, Methodology, Software, Validation, Investigation, Data curation, Visualization, Writing -- original draft.

\textbf{Qiyu Wu:} Writing -- review \& editing. 

\textbf{Yoshimasa Tsuruoka:} Supervision, Funding acquisition.

%% file: sections/app.tex
\section{Training details}
\label{app:train_details}

\paragraph{Hardware.}
We use 32 computing nodes, each equipped with an NVIDIA GH200 Grace Hopper Superchip.

\paragraph{Software.}
We use PyTorch (2.9.0+cu128) FSDP~\citep{10.14778/3611540.3611569}, transformers (4.57.1), datasets (4.3.0) and accelerate (1.11.0) for training. We write our training code referring to the following codebases: GradCache\footnote{\url{https://github.com/luyug/GradCache}}~\citep{gao-etal-2021-scaling}, GritLM\footnote{\url{https://github.com/ContextualAI/gritlm}}~\citep{muennighoff2025generative} and the HuggingFace transformers (4.57.1)\footnote{\url{https://github.com/huggingface/transformers/tree/v4.57.1}} trainer~\citep{wolf-etal-2020-transformers}.


The hyperparameters and implementation details about \methodname{} are described in Table~\ref{tab:stage1_hyperparameters_v1}.

The consumed GPU device memory will explode if we directly use the training loss $\mathcal{L}$ to train LLMs, even smaller ones like 1.7B with eight A100s 80GB because of the computation of contrastive learning and our training examples are relative long due to the incorporation of reasoning traces. To solve the problem, we utilize GradCache~\citep{gao-etal-2021-scaling} for $\mathcal{L}_{\text{Rep}}$ and gradient accumulation for $\mathcal{L}_{\text{Gen}}$ and $\mathcal{L}_{\text{Recons}}$ during training.

\begin{table}[ht]
\centering
\caption{Hyperparameters and hardware accelerators for training \methodname{} models.}
    \begin{tabular}{@{}lll@{}}
    \toprule
    \textbf{Hyperparameter}                                                       & \textbf{\methodname{}-1.7B}        & \textbf{\methodname{}-4B}         \\ \midrule
    Batch size (per device)                                                       & 8                        & 8                \\
    Num of register tokens per sequence                                           & 128                      & 128              \\
    Run / total training steps                                                    & 550 / 2483               & 440/2016         \\
    Total training size                                                           & 635,752                  & 516,336    \\
    Warm-up ratio                                                                 & 0                        & 0 \%             \\
    Learning rate                                                                 & 5e-5                     & 5e-5             \\
    Text embedding pooling                                                        & mean pooling             & mean pooling  \\
    Num of positives                                                              & 1                        & 1                \\
    Num of negatives                                                              & 1                        & 1                \\
    Optimizer                                                                     & adamw\_torch             & adamw\_torch     \\
    Mixed precision training                                                      & bf16                     & bf16             \\
    Temperature for InfoNCE loss                                                  & 0.02                     & 0.02             \\
    Maximum sequence length                                                       & 14,336                   & 10,240           \\
    Base model                                                                    & Qwen3-1.7B-Base          & Qwen3-4B-Base    \\
    Hardware accelerator                                                          & NVIDIA GH200             & NVIDIA GH200 \\
    Num of GPUs                                                                   & 32                       & 32                \\
    Num of GPU hours                                                              & 32 * 40                  & 32 * 40           \\
    \bottomrule
    \end{tabular}%
\label{tab:stage1_hyperparameters_v1}
\end{table}

\paragraph{Training Data.}
\label{app:training_data}

The sources of training data are listed as follows:
\begin{itemize}
    \item \textbf{General Reasoning}: GeneralThought\footnote{\url{https://huggingface.co/datasets/RJT1990/GeneralThoughtArchive}} and 0.9M DeepSeek-R1 reasoning traces~\citep{zhao20251}.
    \item \textbf{Math Reasoning}: OpenMathReasoning~\citep{moshkov2025aimo}.
    \item \textbf{Reasoning-enhanced Retrieval}: ReasonEmbed~\citep{chen2025reasonembed}.
    \item \textbf{High Reasoning Coverage Agentic Data}: DR-Tulu-SFT~\citep{shao2025dr}, ADP Dataset V1~\citep{song2025agent}.
\end{itemize}
We randomly sample examples from the above dataset as our training data.
For the 1.7B model, the total number of training data is 635,752.
For the 4B model, we sampled 516,336 training examples.
Note that our models were trained on only a subset of the sampled training dataset as shown in Table~\ref{tab:stage1_hyperparameters_v1}.

    

\begin{table}
    \centering
    \caption{Reconstruction prompt set.}
    \label{app_table_recons_prompt_set}
    \begin{promptbox}[Reconstruction prompts]
    "What was discussed in the previous conversation?",
    
    "What did we talk about in the last conversation?",
    
    "What did we cover in our last discussion?",
    
    "What were the key points of the previous conversation?",
    
    "Can you give an overview of what we talked about earlier?",
    
    "Remind me what our last conversation was about.",
    
    "What were we discussing earlier?",
    
    "What did we go over in the previous chat?"
    
    \end{promptbox}
\end{table}

\section{Evaluation details}
\label{app:eval_details}
\paragraph{Text retrieval.}
BRIGHT~\citep{su2025bright} is a comprehensive and difficult text retrieval benchmark covering 12 domains/datasets where queries usually require intensive reasoning to find relevant documents. Thus it is utilized to evaluate the text retrieval performance.
BRIGHT benchmark (version) evaluation is conducted on the same server with 8 NVIDIA A100 GPUs with 80 GB memory.

\begin{table}[htbp]
\caption{{Models benchmarked in experiments. This table is from the BRIGHT paper~\citep{su2025bright}.} 
}
\centering
\resizebox{\textwidth}{!}{
\begin{tabular}{l|ccccccc}
    \toprule
     & \textbf{Size} & \textbf{Architecture} & \textbf{Max $|Q|$} & \textbf{Max $|D|$} & \textbf{Instruction} & \textbf{Version} & \textbf{License} \\
    \midrule
    \multicolumn{7}{c}{Sparse model} \\
    \midrule
    BM25 & N/A & Sparse & $\infty$ & $\infty$ & No & gensim\footnote{https://github.com/piskvorky/gensim} & LGPL-2.1-only\\
    \midrule
    \multicolumn{7}{c}{\textit{Open-sourced models (<1B)}} \\
    \midrule
    SBERT & 109M & Encoder & 512 & 512 & No & all-mpnet-base-v2 & Apache-2.0 \\
    BGE & 335M & Encoder & 512 & 512 & No & bge-large-en-v1.5 & MIT\\
    Inst-L & 335M & Encoder & 2048 & 2048 & Yes & instructor-large & Apache-2.0 \\
    \midrule
    \multicolumn{7}{c}{\textit{Open-sourced models (>1B)}} \\
    \midrule
    Inst-XL & 1.5B & Encoder & 2048 & 2048 & Yes & instructor-xl & Apache-2.0 \\
    E5 & 7.1B & Decoder & 4096 & 4096 & Yes & e5-mistral-7b-instruct & MIT \\
    GritLM & 7.1B & Decoder & 256 & 2048 & Yes & GritLM-7B & Apache-2.0 \\
    SFR & 7.1B & Decoder & 4096 & 4096 & Yes & SFR-Embedding-Mistral & CC-BY-NC-4.0\\
    Qwen & 7.7B & Decoder & 8192 & 8192 & Yes & gte-Qwen1.5-7B-instruct & Apache-2.0 \\
    \midrule
    \multicolumn{7}{c}{\textit{Proprietary models}} \\
    \midrule
    Cohere & N/A & Dense & 512 & 512 & No & Cohere-embed-english-v3.0 & Company\\
    \multirow{2}{*}{\centering Google}& \multirow{2}{*}{\centering 1.2B}& \multirow{2}{*}{\centering Dense}& \multirow{2}{*}{\centering 2000}& \multirow{2}{*}{\centering 2000}& \multirow{2}{*}{\centering Yes}& \multirow{2}{10em}{\centering text-embedding-preview-0409, dimension=768 } & \multirow{2}{*}{\centering Company}\\ \\
    OpenAI & N/A & Dense & 8191 & 8191 & No & text-embedding-3-large & Company\\
    Voyage & N/A & Dense & 16000 & 16000 & Yes & voyage-large-2-instruct & Company\\
    \bottomrule
\end{tabular}
}
\label{tab:models}
\end{table}

\paragraph{Document compression.}
Document compression evaluation is conducted on a server with 2 NVIDIA RTX PRO 6000 GPUs.

\begin{table}[h]
    \centering
    \caption{Statistics of the Wikipedia markdown documents grouped by document-length intervals. Token counts are computed using the GPT-2 tokenizer~\citep{radford2019language}.}
    \label{tab:wiki-length-stats}
    \begin{tabular}{ll}
    \toprule
        \textbf{Length range} & \textbf{\# Documents} \\
    \midrule
        $[0,\ 128)$     & 962  \\
        $[128,\ 256)$   & 755  \\
        $[256,\ 512)$   & 719  \\
        $[512,\ 1024)$  & 333  \\
    \midrule
        $[0,\ \infty)$  & {2952} \\
    \bottomrule
    \end{tabular}
\end{table}


\paragraph{Generative tasks.}

Generative task evaluation is conducted on a server with 2 NVIDIA RTX PRO 6000 GPUs.

\paragraph{RAG.}
RAG evaluation is conducted on a server with 8 NVIDIA A100 GPUs with 80GB device memory.

\paragraph{Hybrid paged attention.} The latency testing of hybrid paged attention is also conducted on a server with one NVIDIA A100 GPU with 80GB device memory.

\paragraph{Prompt templates.}

Table~\ref{tab:query_prompt},~\ref{tab:doc_prompt},~\ref{tab:qa_prompt},~\ref{tab:rag_prompt} and~\ref{tab:grit_rag_prompt} present various prompt templates used in our experiments.

\begin{table}
    \centering
    \caption{\methodname{} query prompt used to extract query embeddings.}
    \label{tab:query_prompt}
    \begin{promptbox}[Query prompt]
    Given a question, your mission is to follow the instructions below:
    
    1. Identify the essential problem.
    
    2. Think step by step to reason and describe what information could be relevant and helpful to address the questions in detail.
    
    3. Draft an answer with as many thoughts as you have.
    
    The given question:
    \{query\}
    \end{promptbox}
\end{table}

\begin{table}
    \centering
    \caption{\methodname{} document prompt.}
    \label{tab:doc_prompt}
    \begin{promptbox}[Document prompt]
    Represent this text: "\{document\}"
    \end{promptbox}
\end{table}

\begin{table}
    \centering
    \caption{\methodname{} QA prompt.}
    \label{tab:qa_prompt}
    \begin{promptbox}[QA prompt]
    To answer the following question, you first think about the reasoning process and then provide the user with the answer. The reasoning process and answer are enclosed within <think> and <answer> tags, respectively, i.e., <think> reasoning process here </think><answer> answer here </answer>.
    
    Question: \{user\_question\}
    \end{promptbox}
\end{table}

\begin{table}
    \centering
    \caption{\methodname{} RAG prompt (regular document).}
    \label{tab:rag_prompt}
    \begin{promptbox}[\methodname{} RAG prompt]
    <|im\_start|>user
    
    \{Documents\}
    
    To answer the following question, you first think about the reasoning process and then provide the user with the answer. The reasoning process and answer are enclosed within <think> and <answer> tags, respectively, i.e., <think> reasoning process here </think><answer> answer here </answer>.

    Question: \{user\_question\} 
    
    <|im\_end|>
    
    <|im\_start|>assistant
    
    \end{promptbox}
\end{table}

\begin{table}
    \centering
    \caption{\grit{} RAG prompt.}
    \label{tab:grit_rag_prompt}
    \begin{promptbox}[\grit{} RAG prompt]
    <|user|>
    
    \{title\} \{text\}
    
    \{query\}
    
    Answer the prior query while optionally using the context prior to it
    
    <|assistant|>

    The answer is
    \end{promptbox}
\end{table}


\paragraph{Latency testing details.}
\label{app:hpa_hyperparameters}

For the hybrid paged attention-based LLM serving system, the hyperparameters are shown in Table~\ref{tab:hpa_hyperparameters}.

\begin{table}[h]
    \centering
    \caption{Hyperparameters used in hybrid paged attention-based LLM serving system when conducting latency testing experiments.}
    \label{tab:hpa_hyperparameters}
    \begin{tabular}{lc}
    \toprule
        \textbf{Hyperparameter} & \textbf{Value}  \\
    \midrule
         tensor\_parallel\_size & 1 \\
         gpu\_utilization & 0.9 \\
         temperature & 0.1 \\
         max\_num\_seqs & 16 \\
         max\_num\_batched\_tokens & 16000 * 16 \\
    \bottomrule
    \end{tabular}
\end{table}

The detailed actual inference time of the latency testing experiments is shown in Table~\ref{tab:detailed_actual_inference_time}.
\begin{table}[h]
    \centering
    \caption{Comparison of averaged actual inference time (s) per user query between the naive and HPA-based inference implementation. The batch size is set to 1 for naive implementation. The hyperparameters of HPA-based inference implementation are shown in Table~\ref{tab:hpa_hyperparameters}.}
    \label{tab:detailed_actual_inference_time}
    \begin{tabular}{lccccc}
        \toprule
        \multirow{2.5}{*}{\textbf{Implementation}} & \multicolumn{4}{c}{\textbf{Max new tokens}} & \multirow{2.5}{*}{\textbf{Test dataset (number)}} \\
        \cmidrule(lr){2-5}
         & 128 & 256 & 512 & 1024 \\
        \midrule
        \multicolumn{6}{l}{\textit{Regular Gen. / Query Embedding}} \\
        \midrule
        \methodname{}-1.7B-Naive &  7.50 & 10.97 & 18.75 & 32.36 & \multirow{4}{*}{{BRIGHT Queries (200)}}\\
        \methodname{}-1.7B-HPA   & 0.64 & 0.95 & 1.59 & 2.80 \\
        \methodname{}-4B-Naive   & 9.59 & 14.53 & 23.91 & 41.18 \\
        \methodname{}-4B-HPA     & 0.81 & 1.23 & 2.04 & 3.59 \\
        \midrule
        \multicolumn{6}{l}{\textit{Document Embedding}} \\
        \midrule
        \methodname{}-1.7B-Naive & 3.77 & 3.81 & 3.80 & 3.84 & \multirow{4}{*}{{PwC Documents (200)}}\\
        \methodname{}-1.7B-HPA & 0.33 & 0.34 & 0.34 & 0.33 \\
        \methodname{}-4B-Naive & 4.93 & 5.01 & 5.02 & 5.06 \\
        \methodname{}-4B-HPA & 0.43 & 0.43 & 0.42 & 0.43 \\
        \midrule
        \multicolumn{6}{l}{\textit{Regular RAG (regular document)}} \\
        \midrule
        \methodname{}-1.7B-Naive & 7.63 & 11.47 & 18.14 & 24.92 & \multirow{4}{*}{{PwC Queries and Documents (200)}}\\
        \methodname{}-1.7B-HPA & 0.65 & 0.97 & 1.57 & 2.21 \\
        \methodname{}-4B-Naive & 9.69 & 14.49 & 23.02 & 31.36 \\
        \methodname{}-4B-HPA   & 0.83 & 1.24 & 1.99 & 2.62 \\
        \midrule
        \multicolumn{6}{l}{\textit{Latent Memory-Augmented Generation (compressed document)}} \\
        \midrule
        \methodname{}-1.7B-Naive & 6.17 & 9.90 & 17.54 & 31.25 & \multirow{4}{*}{{PwC Queries and Documents (200)}}\\
        \methodname{}-1.7B-HPA & 0.32 & 0.63 & 1.25 & 2.49 \\
        \methodname{}-4B-Naive & 7.92 & 12.67 & 22.56 & 39.45   \\
        \methodname{}-4B-HPA   & 0.41 & 0.81 & 1.59 & 3.22 \\
        
        \bottomrule
    \end{tabular}
\end{table}

\paragraph{Naive inference implementation.}
\label{app:naive}
We adapted the huggingface generation implementation to fit the needs of \methodname{} models.
Specifically, the first step is to generate tokens like regular LLMs. After that, the second step is to use \tokens{} as the input to obtain their KV cache (compressed KV cache of the context) and text embedding via the mean pooling operation on the last hidden states.
The optional final step is to augment the next user query with the compressed KV cache and the position ids to conduct conditional generation, that is, latent memory-augmented generation.

We set the batch size to 1 for all cases when using the naive inference implementation.
We run naive inference implementation on one NVIDIA A100 GPU.

The naive implementation of regular generation, RAG, query embedding and document embedding is shown in Figure~\ref{lst:algorithm}:
\begin{figure}[h]
\begin{lstlisting}[language=Python, numbers=left, firstline=19, lastline=75]

class GRCGenerateMixin:
    @torch.no_grad()
    def generate_with_compress(
        self,
        input_ids,
        max_new_tokens=50,
        max_length=None,             
        temperature=1.0,
        top_k=0,
        top_p=1.0,
        repetition_penalty=1.0,
        logits_processor=None,
        stopping_criteria=None,
        pad_token_id=None,
        eos_token_id=None,
        use_cache=True,
        **model_kwargs,
    ):
        batch_size, initial_len = input_ids.shape
        
        out = self.generate_with_kv(
            input_ids=input_ids,
            max_new_tokens=max_new_tokens,
            max_length=max_length,             
            temperature=temperature,
            top_k=top_k,
            top_p=top_p,
            repetition_penalty=repetition_penalty,
            logits_processor=logits_processor,
            stopping_criteria=stopping_criteria,
            pad_token_id=pad_token_id,
            eos_token_id=eos_token_id,
            use_cache=use_cache,
        )
        generated_ids = out["sequences"]
        
        past, last_hidden_state = self.append_register_tokens_with_kv(
            batch_size=batch_size,
            past_key_values=out["past_key_values"],
            eos_token_id=eos_token_id,
        )
        raw_reps = last_hidden_state
        
        embedded_raw_reps= self.embed_adapter(raw_reps)
        
        if self.pooling_method == "lasttoken":
            embedding = embedded_raw_reps[:, -1, :]
        elif self.pooling_method == "register_token_mean":
            s = torch.sum(embedded_raw_reps, dim=1)
            embedding = s / raw_reps.shape[1]
        
        if self.normalized:
            embedding = torch.nn.functional.normalize(embedding, p=2, dim=-1)
        
        return generated_ids, raw_reps, embedding, past
        
\end{lstlisting}
    \caption{Naive inference implementation.}
    \label{lst:algorithm}
\end{figure}

\paragraph{Hybrid paged attention implementation.}
Our hybrid paged attention implementation builds upon nano-vllm\footnote{\url{https://github.com/GeeeekExplorer/nano-vllm}}, a compact and efficient paged attention implementation.